\documentclass[runningheads]{llncs}
\usepackage{graphicx}
\usepackage{comment}
\usepackage{amsmath,amssymb} 
\usepackage{color}
\usepackage{times}
\usepackage{epsfig}
\usepackage{amsmath}
\usepackage{amssymb}
\usepackage{caption}

\usepackage{subfigure,times,multirow}
\usepackage{array}
\usepackage[dvipsnames]{xcolor}
\usepackage[ruled]{algorithm2e}
\usepackage{multirow}
\usepackage{cite} 
\usepackage{kotex}
\usepackage[normalem]{ulem} 
\usepackage[keeplastbox]{flushend}
\usepackage[verbose=silent]{microtype}
\usepackage{soul}
\usepackage[export]{adjustbox}
\usepackage{color}
\usepackage{kotex}

\newcommand{\jy}[1]{{\color{black}{#1}}}
\newcommand{\nj}[1]{{\color{black}{#1}}}
\newcommand{\hj}[1]{{\color{black}{#1}}}

\begin{document}
\pagestyle{headings}
\mainmatter
\def\ECCVSubNumber{1491}  

\title{SeqHAND: RGB-Sequence-Based\\3D Hand Pose and Shape Estimation} 

\titlerunning{SeqHAND}
%
\author{
John Yang\inst{1}\and 
Hyung Jin Chang\inst{2}\and 
Seungeui Lee\inst{1}\and 
Nojun Kwak\inst{1}\thanks{Corresponding Author}
}
\authorrunning{J. Yang et al. }
%
\institute{Seoul National University, Seoul, South Korea \email{\{yjohn,seungeui.lee,nojunk\}@snu.ac.kr}
\and University of Birmingham, Birmingham, UK\\
\email{h.j.chang@bham.ac.uk}
}
\maketitle

\begin{abstract}
3D hand pose estimation based on RGB images has been studied for a long time. 
Most of the studies, however, have performed frame-by-frame estimation based on independent static images. 
In this paper, we attempt to not only consider the appearance of a hand but incorporate the temporal movement information of a hand in motion into the learning framework
for better 3D hand pose estimation performance, which leads to the necessity of a large scale dataset with sequential RGB hand images.
We propose a novel method that generates a synthetic dataset that mimics natural human hand movements by re-engineering annotations of an extant static hand pose dataset into \textit{pose-flows}.
With the generated dataset, 
we train a newly proposed recurrent framework, exploiting visuo-temporal features from sequential images of synthetic hands in motion and emphasizing temporal smoothness of estimations with a temporal consistency constraint. 
Our novel training strategy of detaching the recurrent layer of the framework during domain finetuning from synthetic to real allows preservation of the visuo-temporal features learned from sequential synthetic hand images.
Hand poses that are sequentially estimated consequently produce natural and smooth hand movements which lead to more robust estimations.
We show that utilizing temporal information for 3D hand pose estimation significantly enhances general pose estimations by outperforming state-of-the-art methods in experiments on hand pose estimation benchmarks.
\end{abstract}

\section{Introduction}
Since expressions of hands reflect much of human behavioral features in a daily basis, 
hand pose estimations are essential for many human-computer interactions,  
such as augmented reality (AR), virtual reality (VR) \cite{7014300} and computer vision tasks that require gesture tracking \cite{CHANG201687}. 
Hand pose estimations conventionally struggle from an extensive space of pose articulations and occlusions including self-occlusions. 
Some recent 3D hand pose estimators that take sequential depth image frames as inputs have tried to enhance their performance considering temporal information of hand motions \cite{hu2019crnn, wu2018context, oberweger2016efficiently, madadi2018top}.
Motion context provides temporal features for narrower search space, hand personalizing, robustness to occlusion and refinement of estimations.
In this paper we focus on the hand pose estimation considering its movements using only RGB image sequences for better inference of 3D spatial information. 

Although the problem of estimating a hand pose in a single RGB image is an ill-posed problem, its performance is rapidly improving due to the development of various deep learning networks \cite{baseline,mueller2018ganerated,ge20193d}.  However, most studies have focused on accurately estimating 3D joint locations for each image without considering motion tendency. 
Pose of hands changes very quickly and in many cases contains more information on the movements of the successive poses than on the momentary ones. 
In addition, the current pose is greatly affected by the pose from the previous frames. 
Until now, there has been 
a lack of research on the estimation network considering the continuous changes of poses.
The main reason that conventional RGB-based deep 3D hand pose estimators \cite{baseline,zhang2019end,baek2019pushing, mueller2018ganerated} have only proposed frameworks with per-frame pose estimation approaches is that any large scale RGB sequential hand image dataset has not been available \nj{unlike the} datasets with static images of hand poses.
The diversity and the authenticity of hand motions along with generalization over skin colors, backgrounds and occlusions is a challenging factor for a dataset to be assured.

\begin{figure}[t]

\begin{minipage}{.49\textwidth}
\centering
\includegraphics[width=\linewidth]{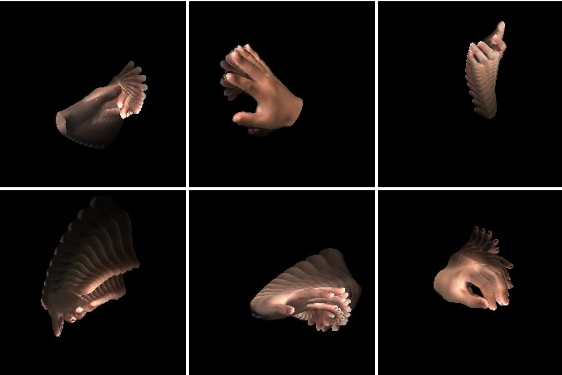}
\vspace{-3mm}
\caption{Illustrations of sequential 2D images of hand pose-flows that are generated by the proposed method.}
\label{fig:pose_flow}
\end{minipage}
\hfill
\begin{minipage}{.49\textwidth}
\centering
\includegraphics[width=\linewidth]{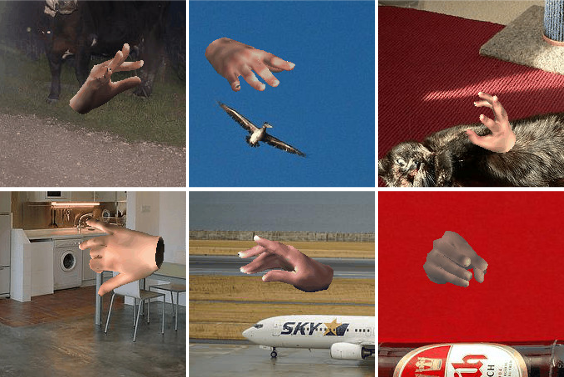}
\vspace{-3mm}
\caption{Each frame of sequential hand motion videos is composed of varying poses and moving backgrounds.}
\label{fig:Data_samples}
\end{minipage}
\vspace{-5mm}
\end{figure}



In this paper, we present a novel perspective on hand pose and shape estimation tasks and propose to consider temporal movements of hands as well as their appearances for more accurate 3D estimations of hand poses based on RGB image inputs.
In order to train a framework that exploits visuo-temporal features to manage successive hand pose images, 
we are required to have sufficient pose data samples that are sequentially correlated. 
We thus propose a new generation method of dataset, SeqHAND dataset, with sequential synthetic RGB images of natural hand movements, re-enginerring extant static hand pose annotations of BigHand2.2M dataset \cite{yuan2017bighand2}.
To effectively test our generated dataset, we extend the framework of \cite{baseline} with a recurrent layer based on empirical validity of its structure. 
Also since it is widely accepted that models trained with synthetic images perform poorly on real images \cite{mueller2018ganerated},
we present a new training pipeline to preserve pre-trained image-level temporal mapping during synthetic-real domain transition. 
\nj{Our contributions to this end} are as follows : 
\vspace{-2mm}
\begin{itemize}
    \item We design a new generation method for sequential RGB image dataset with realistic hand motions that allows 3D hand pose and shape estimators to learn the dynamics of hand pose variations (See Figure \ref{fig:pose_flow}) by proposing a pose-flow generation procedure.
    \item We propose a new recurrent framework with convolution-LSTM layer to directly exploit visuo-temporal information from hand pose and shape variations in image space and map to 3D space. 
    \item We present a novel training pipeline of preserving extracted spatio-temporal features from sequential RGB hand images during domain finetuning from synthetic to real.
    \item Our approach achieves not only state-of-the-art performance in standard 3D hand pose estimation dataset benchmarks, but also smooth human-like 3D pose fittings for the image sequences.
\end{itemize}
\vspace{-2mm}
To the best of our knowledge, we propose the first deep-learning based 3D hand pose and shape estimator without any external 2D pose estimator 
that exploits temporal information directly from sequential RGB images.

\vspace{-2mm}
\section{Related Works}
\vspace{-2mm}
Many approaches \nj{of hand pose estimation} 
(HPE) have been actively studied. 
To acquire hand information, 
the literature of single hand 3D pose estimation has been mainly \nj{based on} 
visual inputs of depth sensors and/or RGB cameras. 

\textbf{Per-frame RGB-based 3D HPE.}
As views of a single 3D scene in multiple perspectives are correlated,
efforts of 3D estimation based on multiple RGB images of a hand have also been introduced \cite{sridhar2014real,simon2017hand,gomez2017large,oikonomidis2011full,de2006regression}.
Multi-view camera setups allow refinements against occlusions, segmentation enhancements and better sense of depth. In the work of \cite{simon2017hand}, bootstrapping pose estimations among images from multiple perspectives help the estimator to retrain badly annotated data samples and refine against occlusions. 
A pair of stereo images provides similar effects in a more limited setting. 
Integration of paired stereo images has yielded better 3D hand pose estimations through manipulations of disparity between paired images \cite{panteleris2017back, rosales20013d, zhang2017hand_stb_stereo, remilekun2017hand}.


Monocular RGB-only setup is even more challenging because it only provides visual 2D vision of hand poses. With deep learning methods that have allowed successful achievements of hand detection \cite{resnet, hand_detection}, deep pose estimators have recently been able to concentrate on per-frame hand 3D pose estimation problems \cite{zimmermann2017learning}.
To overcome the lack of 3D spatial information from the 2D inputs, there are needs of constraints and guidance to infer 3D hand postures \cite{panteleris2018using}.
Most recently, works of \cite{baseline,zhang2019end,baek2019pushing} employ a prior hand model of MANO \cite{MANO} and have achieved significant performance improvement in \nj{the} RGB-only \nj{setup}.

\textbf{Temporal information in 3D HPE.}
Considering temporal features of depth maps, sequential data of hand pose depth images \cite{zhang2017hand_stb_stereo,yuan2017bighand2,mueller2017real,oberweger2016efficiently} have been trained with hand pose estimators. 
The temporal features of hand pose variations are used for encoding temporal variations of hand poses with recurrent structure of a model \cite{hu2019crnn, wu2018context}, modeling of hand shape space \cite{khamis2015learning}, and refinement of current estimations \cite{oberweger2016efficiently, madadi2018top}.
With sequential monocular RGB-D inputs, Taylor et al. \cite{taylor2014user} optimize surface hand shape models, updating subdivision surfaces on corresponding 3D hand geometric models.
Temporal feature exploitation has not been done for deep-learning  based 3D hand pose estimators that take color images as inputs because 
\nj{large scale sequential RGB hand pose datasets have not} been available in the literature.
We share the essential motivation with the work of \cite{cai2019exploiting}, but believe that, \nj{even} without \nj{the} assistance of 2D pose estimation results, sequential RGB images provide sufficient temporal information and spatial constraints for better 3D hand pose inference with robustness to occlusions.

\textbf{Synthetic hand data generations.} 
Since RGB images also consist of background noise and color diversity of hands that distract pose estimations, 
synthetic RGB data samples are generated from the hand model to incite the robustness of models \cite{cai2018weakly,baseline,ge20193d,mueller2017real}.
In \cite{spurr2018cross, yang2019disentangling}, cross-modal data is embedded in a latent space, which allows 3D pose labeling of unlabeled samples generated from (disentangled) latent factor traverses. 
Mueller et al. \cite{mueller2018ganerated} had applied cycleGAN \cite{zhu2017unpaired} 
for realistic appearances of generated synthetic samples to reduce the synthetic-real domain gap.
While there have been recent attempts to solve an issue of lacking reliable RGB datasets through generations of hand images \cite{baseline, zimmermann2017learning, mueller2018ganerated, cai2018weakly, spurr2018cross},
most of the works have focused on generation of realistic appearances of hands that are not in motions.
To strictly imitate human perception of hand poses,
it is critical for RGB-based hand pose estimators to understand the dynamics of pose variations in a spatio-temporal space. 
We further consider that synthetic hand pose dataset in realistic motions provides efficient information for pose estimations as much as appearances.


\vspace{-2mm}
\section{SeqHAND Dataset}
\vspace{-3mm}
\subsection{Preliminary: MANO Hand Model}
\vspace{-2mm}
MANO hand model \cite{MANO} is a mesh deformation model that takes two low-dimensional parameters $\theta$ and $\beta$ as inputs for controlling the pose and the shape, respectively, of the 3D hand mesh outputs. 
With a given mean template $\bar{T}$, the rigid hand mesh is defined as:
\begin{equation}
    M(\theta, \beta) = W(T(\theta, \beta, \bar{T}), J(\beta), \theta, \omega)
\end{equation}
where $T(\cdot)$ defines the overall shape for the mesh model based on pre-defined deformation criteria with pose and shape, and
$J(\cdot)$ yields 3D joint locations using a kinematic tree.
$W(\cdot)$ represents the linear blend skinning function that is applied with blend weights $\omega$.
MANO model may take up to 45-dimensional pose parameters $\theta$ and 10-dimensional shape parameters $\beta$ while the original MANO framework uses 6-dimensional PCA (principal component analysis) subspace of $\theta$ for computational efficiency.

\noindent \textbf{2D Reprojeciton of MANO Hands: }
The location of joints $J(\beta)$ can be globally rotated based on the pose $\theta$, denoted as $R_\theta$,
to obtain a hand posture $P$ with corresponding 3D coordinates of 21 joints:
\begin{equation}
    P = J(\theta, \beta) = R_\theta(J(\beta)).
\end{equation}

After 3D estimations for mesh vertices $M(\theta, \beta)$ and joints $J(\theta, \beta)$ are computed by MANO model, in \cite{baseline}, 
3D estimations are re-projected to 2D image plane with a weak-perspective camera model to acquire 2D estimations with a given rotation matrix $R \in SO(3)$, a translation $t \in \mathbb{R}^2$ and a scaling factor $s \in \mathbb{R}^+$ : 
\begin{align}
    M_{2D} &= s\Pi R M(\theta, \beta) + t
    \label{eq:2D_1}\\
    J_{2D} &= s\Pi R J(\theta, \beta) + t 
    \label{eq:2D_2}
\end{align}
where $\Pi$ represents orthographic projections. 
Hand mesh $M(\theta, \beta)$ is composed of 1,538 mesh faces and defined by 3D coordinates of 778 vertices, and joint locations $J(\theta, \beta)$ are represented by 3D coordinates of 21 joints. 
The re-projected 2D coordinates of $M_{2D}$ and $J_{2D}$ are represented in 2D locations in the image coordinates.
We have utilized MANO hand model in both synthetic hand motion data generation and the \nj{proposed pose and shape estimator}.


\vspace{-3mm}
\subsection{Generation of SeqHAND Dataset}
\vspace{-2mm}
Although the potential of temporal features have been shown promising results for 3D HPE tasks \cite{cai2019exploiting,oberweger2016efficiently,taylor2014user}, large scale RGB sequential hand image datasets have not
been available during recent years in the literature of RGB-based 3D HPE.
In this section, we describe a new generation method of hand motions that consist of sequential RGB frames of synthetic hands.

\begin{figure}[t]
\centering
\begin{minipage}{.49\textwidth}
\centering
\includegraphics[scale=0.65, trim={0cm 0cm 0 0.1cm},clip]{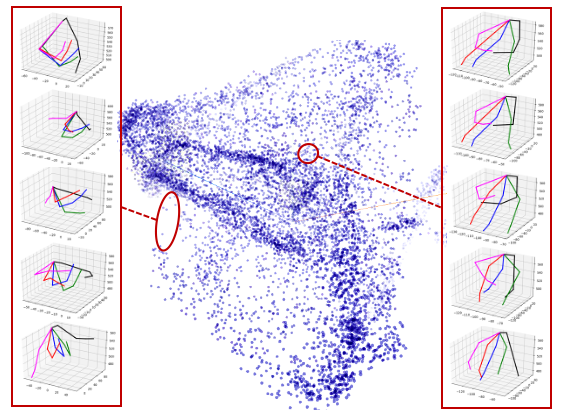}
\vspace{-2mm}
\caption{3D t-SNE visualization of 10,000 of BH data samples randomly selected. BH dataset completes a pose space that covers previously reported datasets, having a dense pool of \nj{related neighboring poses.}}
\label{fig:bh_samples}
\vspace{-4mm}
\end{minipage}
\hfill
\begin{minipage}{.49\textwidth}
\centering
\includegraphics[scale=0.67]{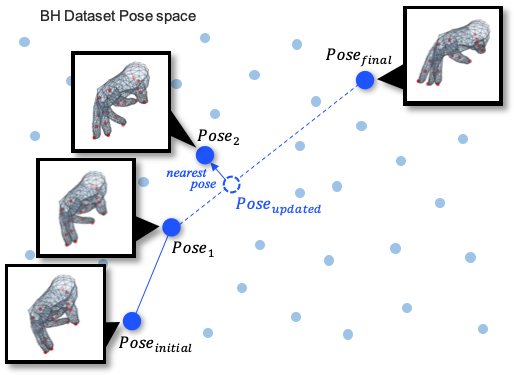}
\vspace{-2mm}
\caption{An illustration of the pose-flow generation procedure. 
All poses per pose-flow are selected from the annotations of BH dataset. 
At each frame, a current pose is updated by the difference between the previous pose and the final pose. 
The pose nearest to the updated pose is then selected for the frame.
}
\label{fig:pose_flow_gen_diagram}
\vspace{-4mm}
\end{minipage}

\end{figure}

To generate sequential RGB image \nj{data} with human-like hand motions, 
all poses during the variation from an initial pose to a final pose need to be realistically natural.
We thus have utilized BigHand2.2M (BH) \cite{yuan2017bighand2} for sequential hand motion image dataset generation.
BH dataset consists of 2.2 million depth maps with 3D annotations for joint locations acquired from 2 hour-long hand motions collected from 10 real subjects. 
With BH datasets, the generated samples are expected to inherit the manifold of its broad real human hand articulation space and kinematics of real hand postures.
\jy{3D mapping of BH samples using t-SNE \cite{maaten2008tsne} in Figure \ref{fig:bh_samples} shows how dense the pose samples are collected.}
Such density of BH dataset with a more complete range of variation than reported datasets is considered sufficient for various pose generations.



\noindent\textbf{Pose-flow generation: } 
A hand motion can be defined as a continuous change from a pose to another during a period of a time. 
We firstly define a \textit{pose-flow}, a set of poses at each time step during the variation.
Putting gradually changing poses in a sequential manner, 
we newly propose a \textit{pose-flow} generation method.
For each pose-flow generation, 
an initial and \nj{a} final poses, $P_{initial}$ and $P_{final}$, are independently and randomly selected from BH dataset. 
While varying from the initial to the final pose during $n$ frames, 
the coordinates of joints are updated by $\alpha/n$ of the difference between the current coordinates and the ones of the final pose.\footnote{
\nj{Note that d}irect random samplings from continuous pose parameter space $\theta \in \mathbb{R}$ does not assure diversity and authenticity of poses \cite{MANO}.} 
The update size $\alpha$ is empirically chosen for the desirable speed of pose variations.
A pose $P_i^{BH}$ from BH dataset that is the nearest to the updated pose in terms of Euclidean distance is then newly selected as the current pose for the $k$-th frame:
\begin{align}
   &P_0=P_{initial}^{BH}\\
    &P_{updated}  = P_{k-1} - \frac{\alpha}{n} (P_{k-1} - P^{BH}_{final})\\
     &P_{k} = P_i^{BH}\quad \text{s.t. }\quad \min_{i}||P_{updated} - P_i^{BH}||.
\end{align}
The overall procedure of the Pose-flow generation is summarized in Figure \ref{fig:pose_flow_gen_diagram}.
\jy{\nj{The intermediate pose (\textit{P$_{\text{updated}}$})} is calculated as stochastic update. Such stochasticity of our pose updates helps avoiding strict updates of pose gradients and encourages wandering more within the pose space.
Pose selections from the BH annotations, again, allows assurance on the authencity of hand poses during the variation.}

To generate RGB images for a pose, an encoder with four MLP layers which takes inputs of 3D coordinates for 21 joints of all BH joint annotations is trained to output corresponding pose parameters $\theta$ for MANO hand model based on the reconstruction loss between the inputs and the outputs of the hand model.
During this training, MANO hand model is detached from the training. 
For each pose at a frame, 
we feed corresponding 21 joint location coordinates to the the encoder to acquire a hand mesh model in the desired pose. 
The rendered mesh model is then re-projected to an image plane.
As done in \cite{baseline},
we assign each vertex in a mesh the RGB value of predefined color templates of hands to create appearances of hands.
Sampled hand shape parameter $\beta \in [-2,2]^{10}$ and selected color template are set unchanged along per flow.
Camera parameters of rotation $R$, scale $s$ and translation $t$ factors are independently sampled for starting and ending poses and updated at each frame in the same way as the poses are.
All frames are in the size of $w$ and $h$.
Figure \ref{fig:pose_flow} depicts illustrations of our generated pose-flows. 

Further mimicking images of hand motions in the wild,
we sample two (initial and ending) random patches from VOC2012 data \cite{pascal-voc-2012} with the size of $w$ and $h$ and move the location of the patch for backgrounds along the frames.
As Table \ref{tab:seqhand_position} denotes, the generated SeqHAND dataset provides not only both $3^{rd}$-person and egocentric viewpoints of hand postures but also sequential RGB images of hand poses that firstly allow data-hungry neural networks to exploit visuo-temporal features directly from RGB inputs.\footnote{\nj{Although we can generate as many synthetic data as we want, our SeqHand dataset contains 400K/10K samples used for training/validation.}}

\begin{table}[t]
\centering
\caption{Among the contemporary 3D hand pose datasets, SeqHAND dataset is the first dataset for 3D hand pose estimations that provides sequential RGB hand image frames along with stable annotations in both $3^{rd}$-person and egocentric perspectives. }
\vspace{1mm}
\resizebox{0.85\linewidth}{!}{
\begin{tabular}{|l|c|c|c|c|c|}
\cline{1-6}
                                 {Datasets}&{ RGB/Depth }&{ Real/Synth }&{ Static/Sequential }& { $3^{rd}$/Ego view }& { \# of frames }\\ \hline \hline
\multicolumn{1}{|l|}{SynthHands\cite{mueller2017real}} & RGB+Depth      & Synth      & Static            & Ego          & 63k        \\ \hline
\multicolumn{1}{|l|}{RHD\cite{zimmermann2017learning}}        & RGB+Depth      & Synth      & Static            & $3^{rd}$          & 43.7k      \\ \hline
\multicolumn{1}{|l|}{NYU \cite{tompson2014real}}        & Depth     & Real       & Sequential        & $3^{rd}$          & 80k        \\ \hline
\multicolumn{1}{|l|}{ICVL \cite{tang2014latent}}       & Depth     & Real       & Sequential        & $3^{rd}$          & 332.5k     \\ \hline
\multicolumn{1}{|l|}{FHAD\cite{garcia2018first}}       & Depth     & Real       & Sequential        & Ego          & 100k       \\ \hline
\multicolumn{1}{|l|}{MSRA15\cite{sun2015cascaded}}     & Depth     & Real       & Sequential        & $3^{rd}$          & 76,375     \\ \hline
\multicolumn{1}{|l|}{MSRC\cite{sharp2015accurate}}       & Depth     & Synth      & Sequential        & $3^{rd}$+Ego         & 100k       \\ \hline
\multicolumn{1}{|l|}{SynHand5M\cite{malik2018deephps}}  & Depth     & Synth      & Sequential        & $3^{rd}$          & 5M         \\ \hline
\multicolumn{1}{|l|}{GANerated\cite{mueller2018ganerated}}  & RGB       & Synth      & Static            & Ego          & 330k       \\ \hline
\multicolumn{1}{|l|}{\textbf{SeqHAND}}    & RGB       & Synth      & Sequential        & $3^{rd}$+Ego         & 410k       \\ \hline
\end{tabular}
}
\vspace{-3mm}
\label{tab:seqhand_position}
\end{table}

\begin{figure}[t]
\centering
\includegraphics[scale=0.60, trim={.1cm 0 0 0},clip]{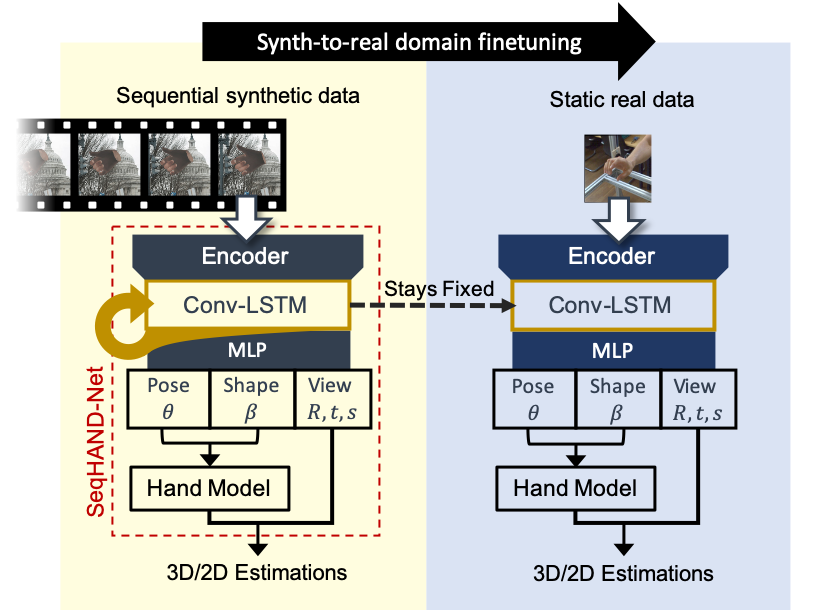}
\caption{Illustration of our proposed training strategy for preservation of temporal features learned from SeqHAND dataset during domain adaptation form synthetic to real. The structure of SeqHAND-Net is created by extending a pose and shape estimator from \cite{baseline} with an extra visuo-temporal feature exploitation layer from sequential RGB image inputs. During finetuning to the real domain, SeqHAND-Net considers each static real data sample as one-frame-long sequence. The high-level temporal features of SeqHAND-Net are preserved while low-level image feature encoding layers are finetuned.}
\label{fig:framework}
\vspace{-3mm}
\end{figure}

\vspace{-2mm}
\hj{\section{SeqHand-Net for Visuo-Temporal Feature Exploitation}}
\vspace{-2mm}

With SeqHAND dataset, we are able to overcome the scarcity of sequential RGB dataset which limits conventional RGB-based 3D HPE methods from exploiting temporal image features. 
\jy{Motivated by \cite{baseline}, we design sequential hand pose and shape estimation network (SeqHAND-Net). 
On top of the encoder network of \cite{baseline}, we incorporate convolution-LSTM (ConvLSTM) layer \cite{convolutionlstm} to capture sequential relationship between consecutive hand poses.}
Our method does not consider additional hand 2D joint locations as inputs, 
and purely performs 3D hand pose estimation based on sequentially streaming RGB images
in an effort to overcome the dependency on external 2D pose estimators.
We also propose, in this section, a training pipeline for domain adaptation from synthetic to real, adapting low-level features with real hand images while preserving high-level visuo-temporal features of hand motions. 


From each frame, a cropped hand image is fed into SeqHAND-Net as illustrated in Figure \ref{fig:framework}. 
Our problem scope is to better perform hand pose estimations on streaming cropped frames that are unseen by the estimator.
The encoder of our SeqHAND-Net has the backbone structure of ResNet-50\cite{resnet} and expects sequential inputs with $k$ frames, 
and it produces a resultant 26-dim. vector for each frame.
A single ConvLSTM is implemented right before the last layer as a recurrent visual feature extractor so that the dynamics of hand motions are embedded in the highest-level latent space.
Learning of hand motion sequential dynamics in the high-level space is important since low-level visual features are changed with the ConvLSTM layer fixed during finetuning for real hand images.
After the recurrent layer, a simple linear mapping layer from hidden features to the output vector is set.
The encoder's resultant vector consists of parameters for pose $\theta \in \mathbb{R}^{10}$, shape  $\beta \in \mathbb{R}^{10}$, scale $s \in \mathbb{R}^+$, translation $t \in \mathbb{R}^2$ and rotation $r \in \mathbb{R}^{3}$ which turns into a matrix $R\in SO(3)$ through Rodrigues rotation formula for Eqs (\ref{eq:2D_1}) and (\ref{eq:2D_2}).

\textbf{Synth-to-Real Domain Transfer with Preservation of Temporal Features: }
As mentioned earlier, many recent researches have used synthetic hand images for pre-training and finetuned into real domain to overcome the scarcity of real hand images.
While finetuning into real domain may allow faster training convergence, further training with a smaller dataset not only causes overfitting and may result in catastrophic forgetting \cite{kirkpatrick2017overcoming}.
To preserve visuo-temporal features learned from synthetic hand motions of SeqHAND dataset, 
we exclude the ConvLSTM layer of SeqHAND-Net from domain transfer to real hand images, allowing the network to only finetune low-level image features.
Only the `Encoder' and `MLP' layers from Figure~\ref{fig:framework} are finetuned with a real static hand image dataset (e.g. FreiHand \cite{zimmermann2019freihand}). 
SeqHAND-Net is therefore trained, considering each image sample as one-frame-long sequential real hand image during domain transition to real.
%

\textbf{Training Objectives: }
The followings are the types of criteria used for training our proposed framework to consider visuo-temporal features and emphasize the temporal smoothness of estimations :
\begin{itemize}
    \item \textbf{2D joint regression loss.} 
        The re-projected 2D joint loss is represented as : 
        \begin{align}
            L_{2D}^J = || J_{2D} - x_{2D}^J ||_1,
        \end{align}
        where $x_{2D}$ represents the ground-truth 2D locations of hand joints within a frame image. 
        We have used the L1 loss because of inaccuracies in annotations in the training datasets.

    \item\textbf{3D joint regression loss.} 
        The ground-truth joint locations and the ones predicted are regressed to be the same using the following loss: 
        \begin{align}
            L_{3D}^J = ||RJ(\theta, \beta) - x_{3D}^J ||_2^2,
        \end{align}
        where $x_{3D}^J$ represents ground-truth 3D joint coordinates.
        If the training dataset provides ground-truth coordinates of 3D vertex points (e.g. FreiHand dataset), the 3D coordinates of each vertex predicted and the ones of ground-truth is minimized as done for 3D joint loss, based on the following loss:
        \begin{align}
            L_{3D}^M = ||RM(\theta, \beta) - x_{3D}^M ||_2^2
        \end{align}
        where $x_{3D}^M$ represents ground-truth 3D mesh vertex coordinates.

    \item\noindent\textbf{Hand mask fitting loss.} 
        The hand mask loss is proposed in \cite{baseline} to fit the shape and pose predictions in the binary mask of hands in the image plane. 
        This loss ensures predicted coordinates of mesh vertices to be inside of a hand region when re-projected:
        \begin{equation}
            L_{mask} = 1 - \frac{1}{N}\sum_i H(M_{2D}^i), \quad
                  H(x)=\begin{cases}
            1, & \text{if $x$ inside a hand region}.\\
            0, & \text{otherwise}.
          \end{cases}
        \end{equation}
        where $H$ is a hand mask indicator function that tells if vertex point $x$ is inside the hand region or not. The loss represents the percentage of vertices that are outside the region.

    \item\textbf{Temporal consistency loss.} 
        For pre-training on SeqHAND dataset, our method needs to be constrained with temporal consistency to ensure smoothness of pose and shape predictions. Similar to \cite{cai2019exploiting}, we have adopted the temporal consistency loss for smoothness of temporal variation of poses: 
        \begin{align}
            L_{temp} = || \beta_{t-1} - \beta_{t}||_2^2 + \lambda_{temp}^\theta || \theta_{t-1} - \theta_{t} ||_2^2.
        \end{align}
        Considering the fact that all hands in a sequence is the same hand for all the frames, we have set the constraint hyper-parameter $\lambda_{temp}^\theta$ a comparably small number as $0.01$ so that temporal variation of hand shapes per image sequence data to be low while pose variation is less constrained but is still assured of temporal smoothness. \jy{While penalizing current estimations with the previous ones, this loss allows the reduction of search space, sequentially natural 3D hand motion estimations, and hand shape personalization}.
    
    \item\textbf{Camera parameter regression loss.} 
        During training with SeqHAND dataset where all ground-truths for pose, shape and viewpoint parameters \{$\theta, \beta, r, t, s$\} are available, 
        our model is trained with L2-norm loss between predictions and the ground-truth. 
        \begin{align}
            L_{cam} =& \sum_{i \in \{\theta,\beta,r,t,s\}} || \hat{i} - i ||_2^2 
        \end{align}
        where $\hat{i}$ and $i$ respectively refer to predicted and ground-truth parameters for pose, shape and viewpoint.

\end{itemize}

\noindent\textbf{Training Loss for SeqHAND Dataset.}
The criterion for pre-training for sequential synthetic hand motion image dataset is a combination of re-projected 2D joint loss $L_{2D}$, a 3D joint loss $L_{3D}$, a temporal consistency loss $L_{temp}$, a loss for camera parameters $L_{cam}$ and the mask fitting loss $L_{mask}$:
\begin{align}
    L =&\lambda_{2D} L_{2D} + \lambda_{3D} L_{3D} + \lambda_{temp} L_{temp}  + \lambda_{cam} L_{cam} + \lambda_{mask} L_{mask}.
\end{align}

\noindent\textbf{Training Loss for Domain Transition to Real.}
We have utilized datasets of Stereo Benchmark  \cite{zhang2017hand_stb_stereo} and FreiHand \cite{zimmermann2019freihand} for domain transfer of our trained network into real domain. 
The two datasets are differently annotated; STB datasets are only annotated with 2D and 3D joint locations while FreiHand dataset provides hand masks along with the 2D and 3D ground-truths.
The loss of adaptation to real hand images is thus a combination of re-projected a 2D joint loss $L_{2D}$, a 3D joint loss $L_{3D}$, a hand mask loss $L_{mask}$, and a temporal loss $L_{temp}$ :
\begin{align}
    L =&\lambda_{2D} L_{2D} + \lambda_{3D} L_{3D} + \lambda_{temp} L_{temp} + \lambda_{mask} L_{mask}.
\end{align}
We have set the weights as $\lambda_{2D}=5, \lambda_{3D}=100, \lambda_{temp}=100$ with $\lambda_{temp}^\theta=2e^{-4}, \lambda_{cam}=1$ and $\lambda_{mask}=10$ for both pre-training with SeqHAND dataset and domain adaptation to real hand images.

\vspace{-2mm}
\section{Experiments}
\vspace{-2mm}


\noindent\textbf{Datasets for Training.} 
For visuo-temporal feature encodings of sequential RGB hand motion frames, we pre-train SeqHAND-Net with our SeqHAND dataset.
We have generated 40,000 SeqHAND sequence training and 1,000 test samples each of which is 10-frames-long.
All images are generated in size of 224$\times$224 to fit for ResNet-50 input size. 
SeqHAND data samples are exemplified in Figure~\ref{fig:pose_flow} and \ref{fig:Data_samples}.

To finetune SeqHAND-Net for synthetic-real domain gap reduction, we have used STB (Stereo Hand Pose Tracking
Benchmark) \cite{zhang2017hand_stb_stereo} and FR (FreiHand) \cite{zimmermann2019freihand} datasets. 
STB dataset consists of real hand images captured in a sequential manner during 18,000 frames with 6 different lighting conditions and backgrounds.
Each frame image is labeled with 2D and 3D annotations of 21 joints.
Since STB dataset has annotations for joint locations of palm centers instead of wrist, we have interpolated related mesh vertices of MANO hand model to mach the annotation of STB dataset. 
The dataset is divided into training and testing sets as done in \cite{baseline} of which testing set is used for evaluation only.

FR dataset has 130,240 data samples that are made up of 32,560 non-sequential real hand images with four different backgrounds. 
Since the dataset has hands that are centered within the image planes, we have modified each sample by re-positioning the hand randomly within the image for more robust training results.
FR dataset provides MANO-friendly annotations of 21 joint 3D/2D locations along with 778 vertex ground-truth 2D/3D coordinates with hand masks.

We have finetuned the SeqHAND-Net pretrained on SeqHAND dataset with real-hand image datasets mentioned above in a non-sequential manner while conserving hand motion dynamic features detached from further learning.

\noindent\textbf{Datasets for Evaluation.} 
We evaluate various framework structures that consider temporal features on the validation set of SeqHAND dataset for the logical framework choice. 
For the comparison against other state-of-the-art methods,
we have selected standard hand pose estimation datasets of the splitted test set of STB, EgoDexter(ED)\cite{mueller2017real} and Dexter+Obeject(DO)\cite{sridhar2016real} in which there exists temporal relations among data samples since our network requires sequential RGB inputs for fair comparisons.
While STB and DO datasets consist of real hand images in $3^{rd}$-person viewpoints, ED dataset has samples that are in egocentric perspective.
For all datasets, our method is evaluated on every frame of input sequences.


\noindent\textbf{Metrics.}
For evaluation results, we measure the percentage of correct key-points for 3D joint locations (3D-PCK) along with the area under the curve (AUC) of various thresholds.
In addition, we provide average Euclidean distance error for all 2D/3D joint key-points so that more absolute comparisons can be made.

\noindent\textbf{Hand Localizations.} For all experiments, we have used MobileNet+SSD version of hand detection implementation \cite{hand_detection} trained with a hand segmentation dataset \cite{bambach2015lending} for providing sequential cropped hand images to SeqHAND-Net. 
For localized hands with tight bounding rectangular boxes, we choose the longer edge with a length size $l$ and crop the region based on the center point of boxes so that the cropped images have a square ratio with width and height size of $2.2*l$, as done in \cite{baseline}.




\begin{table}[t]
\centering
\caption{Ablation study results of various structures of the framework proposed in \cite{baseline} for sequential inputs. Candidate structures are trained with the generated SeqHAND dataset and evaluated on a separate validation set. Candidate methods include the baseline model, a 3D Convolutional Network (I3D-Encoder), a 3D convolutional network encoder with motion feature extractions (MFNet-Encoder), baseline with an LSTM layer (Encoder+LSTM), and baseline with a ConvLSTM layer (Encoder+ConvLSTM).}
\vspace{1mm}
\resizebox{0.8\linewidth}{!}{
\begin{tabular}{|c|c|c|c|c|c}
\hline
            \multirow{2}{*}{Frameworks}                                                               & \multicolumn{2}{c|}{AUC} & \multicolumn{2}{c|}{Error (px/mm)} & \multicolumn{1}{l|}{\multirow{2}{*}{\# params} }       \\ \cline{2-5} 
                                                                         & 2D          & 3D         & 2D              & 3D               & \multicolumn{1}{l|}{}  \\ \hline
                                                                         \hline
\multicolumn{1}{|l|}{ResNet50-Encoder (baseline) \cite{baseline}}                        &0.855 & 0.979 & 3.44 & 7.85 & \multicolumn{1}{c|}{28.8M}   \\ \hline
\multicolumn{1}{|l|}{ResNet101-Encoder \cite{baseline}}                                  & 0.861           &0.981         & 3.31             & 7.54                & \multicolumn{1}{c|}{47.8M}    \\ \hline
\multicolumn{1}{|l|}{I3D-Encoder \cite{carreira2017quo}} & 0.831 & 0.967 & 4.19  & 9.24  & \multicolumn{1}{c|}{31.5M}        \\ \hline
\multicolumn{1}{|l|}{MFNet-Encoder \cite{lee2018motion}} &  0.818 &0.912 &5.48     &10.54 & \multicolumn{1}{c|}{41.7M}      \\ \hline
\multicolumn{1}{|l|}{ResNet50-Encoder+LSTM} & 0.826& 0.956& 4.64& 9.63& \multicolumn{1}{c|}{39.3M}    \\ \hline
\multicolumn{1}{|l|}{\textbf{ResNet50-Encoder+ConvLSTM}}                 &  \textbf{0.873}& \textbf{0.986}& \textbf{3.17}& \textbf{7.18}& \multicolumn{1}{c|}{43.2M}      \\ \hline
\end{tabular}
}
\vspace{-3mm}
\label{tab:ablation1}
\end{table}

\vspace{-3mm}
\subsection{Ablation Study}
\vspace{-2mm}
\hspace{\parindent}\textbf{Framework Selection: }
To show the logic behind the selection of the proposed framework, 
we evaluate various forms of extended baseline model \cite{baseline} shown in Table~\ref{tab:ablation1} for managing sequential inputs on our newly generated SeqHAND dataset.
The extended versions of baseline encoder (ResNet-50) include the baseline model with a LSTM layer\cite{sundermeyer2012lstm}, the baseline model with a ConvLSTM layer\cite{convolutionlstm}, the baseline encoder with the structure of I3D\cite{carreira2017quo} and the baseline encoder with the structure of MF-Net \cite{lee2018motion}. 
Both I3D-Encoder and MFNet-Encoder represent methods that incorporate sequential inputs with 3D convolutional neural network. 
For I3D-Encoder, we have changed few features from the original form of I3D so that its structure fits into the hand pose estimation task. 
The original backbone structure of I3D with Inception modules have changed into ResNet-50 for a fair comparison. 
MFNet is another examplary 3D convolution network proposed specifically for motion feature extractions. 
Of the candidates, the encoder with ResNet-50 backbone structure with a ConvLSTM layer has performed the best.

\begin{table}[t]
\centering
\caption{Performances of differently (partially) trained models on ED, DO, STB datasets.}
\vspace{1mm}
\resizebox{\linewidth}{!}{
\begin{tabular}{|c|c|c|c|c|c|c|}
\cline{1-7}

 \multirow{2}{*}{Methods}  & \multicolumn{3}{c|}{AUC}     & \multicolumn{3}{c|}{Avg. 3D Error (mm)}      \\ \cline{2-7} 
                                & \multicolumn{1}{c|}{ED}   & \multicolumn{1}{c|}{DO}   & \multicolumn{1}{c|}{STB}  & ED     & DO    & STB  \\ \hline \hline
\multicolumn{1}{|l|}{Encoder + Train(SynthHAND)}                         & 0.350 & 0.095 & 0.140 & 52.11 & 100.84 & 68.86   \\ \hline
\multicolumn{1}{|l|}{Encoder + Train(SynthHAND) + Train(FH + STB)}       & 0.397 & 0.516 &\textbf{ 0.985 }& 49.18 & 33.12 &\textbf{ 9.80 }\\ \hline
\multicolumn{1}{|l|}{Encoder + ConvLSTM + Train(SeqHAND)}                       & 0.373 & 0.151 & 0.121 & 52.18 & 81.51 & 71.10 \\ \hline
\multicolumn{1}{|l|}{Encoder + ConvLSTM + Train(SeqHAND) + Train(FH + STB)} & 0.444 & 0.581 & 0.981 & 40.94 & 29.41 & 9.82 \\ \hline
\multicolumn{1}{|l|}{Encoder + ConvLSTM + Train(SeqHAND) + Train$_C$(FH + STB)} &\textbf{ 0.766 }&\textbf{ 0.843 }& 0.978 &\textbf{17.16 }& \textbf{18.12}& 9.87 \\ \hline
\end{tabular}}
\label{tab:Effects}
\vspace{-5mm}
\end{table}

\begin{table}[t]
    \centering
	\begin{minipage}{0.45\linewidth}
	    \footnotesize
		\centering
        \caption{Average 3D joint distance (mm) to ground-truth for RGB Sequence datasets hand pose benchmarks. }		
		\begin{tabular}{l|c|c|c|}
        \cline{2-4}
                                        & \multicolumn{3}{c|}{Avg. 3D Error (mm)} \\ \cline{2-4} 
                                         & ED           & DO          & STB        \\ \hline \hline
        \multicolumn{1}{|l|}{Our Method}           & \textbf{17.16 }        & \textbf{ 18.12 }       & 9.87       \\ \hline
        \multicolumn{1}{|l|}{Bouk. et al. (RGB)}    & 51.87        & 33.16       &\textbf{ 9.76 }       \\     \hline
        \multicolumn{1}{|l|}{Bouk. et al. (Best)}  & 45.33        & 25.53       &\textbf{ 9.76 }          \\     \hline
        \multicolumn{1}{|l|}{Spurr et al.}         & 56.92        & 40.20       & -          \\     \hline
        \multicolumn{1}{|l|}{Zimmer. et al.}      & 52.77        & 34.75       & -          \\ \hline
        \end{tabular}
        \label{tab:auc_n_3derr}
	\end{minipage}
	\hspace{2mm}
	\begin{minipage}{0.45\linewidth}
		\centering
		\includegraphics[width=55mm, trim={.7cm 0 0 0},clip]{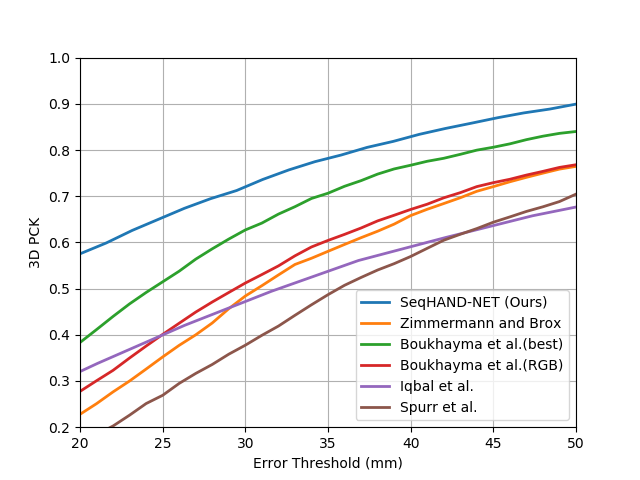}
		\captionof{figure}{3D PCK for ED}
		\label{fig:pck_ed}
	\end{minipage}
	\vspace{-3mm}
	\\
		\begin{minipage}{0.45\linewidth}
		\centering
		\includegraphics[width=55mm, trim={.5cm 0 0 0},clip]{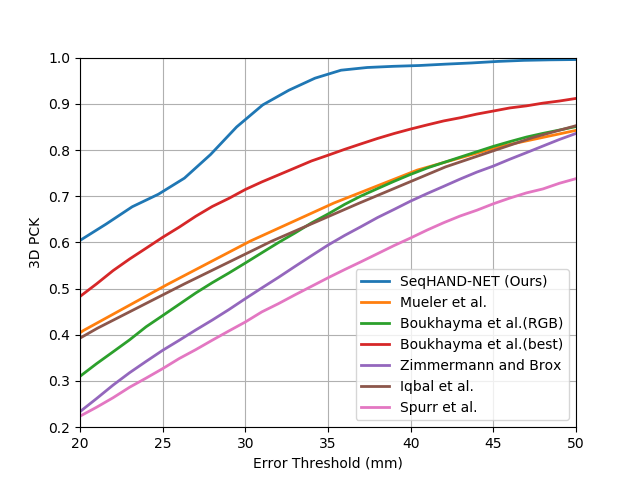}
		\captionof{figure}{3D PCK for DO}
		\label{fig:pck_do}
	\end{minipage}
	\hspace{3mm}
		\begin{minipage}{0.45\linewidth}
		\centering
		\includegraphics[width=55mm, trim={.8cm 0 0 0},clip]{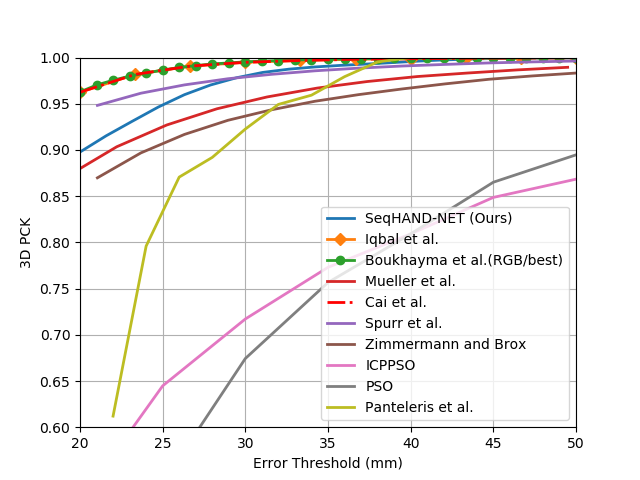}
		\captionof{figure}{3D PCK for STB}
		\label{fig:pck_stb}
	\end{minipage}
	\vspace{-8mm}
\end{table}

\textbf{The Effectiveness of \textit{SeqHAND-Net} and \textit{SeqHAND} Dataset: }
To clarify the effectiveness of our proposed framework and our generated dataset, variations of the proposed method and the baseline model are investigated.  
We report AUCs of 3D PCK curves and average 3D joint location errors for ED, DO and the evaluation set of STB datasets.
In Table \ref{tab:Effects}, `Encoder' denotes the baseline model with ResNet50 backbone structure while `Encoder + ConvLSTM' denotes our proposed framework SeqHAND-Net. 
`Train(SynthHAND)' and `Train(SeqHAND)' represent training a model with synthetic hand image dataset respectively in non-sequential and sequential manner.
`Train(FH + STB)' and `Train$_C$(FH + STB)' refers to training with STB and FreiHand datasets for the synthetic-real domain transition with the ConvLSTM layer, respectively, attached and detached from finetuning with a reasonable computational cost.

We show in the Table \ref{tab:Effects} how much performance enhancement can be obtained with SeqHAND dataset and our proposed domain adaptation strategy.
Encoder with the ConvLSTM layer finetuned to real domain consequently performs similar to the encoder that does not consider visuo-temporal correlations. 
If the ConvLSTM layer is detached from finetuning and visuo-temporal features learned are preserved, the performance significantly improves.
Also, SeqHAND dataset does not consist with any occluded hands except for self-occlusions. 
With training for FH dataset, our method is able to learn the visual features of not only real hands but also occluded real hands since FH dataset's augmentations consist of occlusions. 
Due to the temporal constraint that penalizes large difference among sequential estimations, per-frame estimation performs slightly better for the STB dataset. 

\begin{figure}[!]
\includegraphics[width=\textwidth, trim={.6cm 3cm .6cm 1cm},clip]{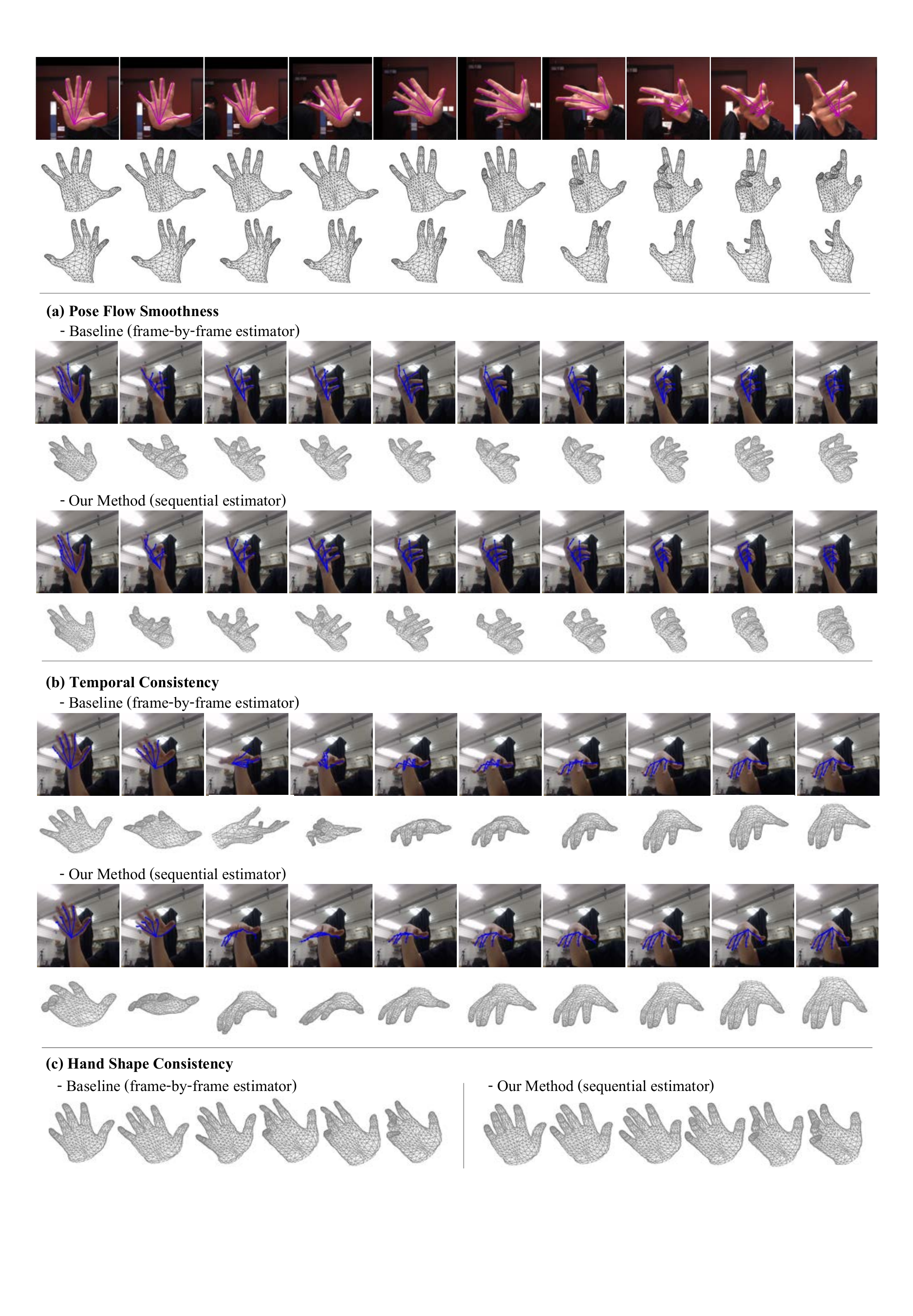}
\centering
\caption{Qualitative Results of SeqHAND-Net.}
\label{fig:qual}
\end{figure}

\vspace{-3mm}
\subsection{Comparison with the State-of-the-art Methods}
\vspace{-2mm}
In Figures~\ref{fig:pck_ed}, \ref{fig:pck_do} and \ref{fig:pck_stb}, we have plotted 3D-PCK graph with various thresholds for STB, ED and DO datasets.
For STB dataset, deep-learning based works of \cite{baseline, cai2018weakly, spurr2018cross, iqbal2018hand, mueller2018ganerated, zimmermann2017learning} and approaches from \cite{panteleris2018using, zhang20163d} are compared. 
\nj{Many} previous methods have reached near the maximum performance for STB dataset.
\nj{With our} temporal constraints and fixing the ConvLSTM layer during finetuning, our method reaches a competitive performance.
For both ED and DO datasets, our method clearly outperforms other methods. 
For ED dataset, contemporary works of \cite{zimmermann2017learning, baseline, iqbal2018hand, spurr2018cross} are compared to our method.
The best performance of the work by Boukhayma et al. \cite{baseline} is reached with inputs of RGB and 2D pose estimations provided by an external 2D pose estimator. 
Our method results in outstanding performance against other compared methods \cite{mueller2018ganerated, zimmermann2017learning, baseline, iqbal2018hand, spurr2018cross} for DO dataset with heavy occlusions, which shows that the learning of pose-flow continuity \nj{enhances robustness} to occlusions.
Temporal information exploitation from sequential RGB images affect our model to be robust against dynamically moving scene.
For more absolute comparisons, we provide our average 3D error of joint location in Table \ref{tab:auc_n_3derr}.


We provide qualitative results in Figure~\ref{fig:qual} for visual comparison against a frame-by-frame 3D pose estimator, our reproduced work of \cite{baseline}. 
All images in the figure are sequentially inputted to both estimators from left to right.
In Figure \ref{fig:qual}(a), per-frame estimations that fit postures at each frame result in unnatural 3D hand posture changes over a sequence while our method's leaning trajectories biased by previous frames produces a natural hand  motion.
When a frame lacks much visual information of hand postures as in the cases of 3rd and 4th frames from the left in Figure \ref{fig:qual}(b), the frame-by-frame estimator's performance significantly decrements. 
As illustrated in Figure \ref{fig:qual}(c), 
our method models the hand shape as consistent as possible per sequence.
During the qualitative evaluation on a RGB image sequence of a single real hand, 
our method's average difference among temporal changes of shape parameters  $\beta_{t-1}-\beta_t$ is 4.16$e^{-11}$ while that of the frame-by-frame estimator is 2.38$e^{-5}$.
The average difference among temporal changes of the pose parameters $\theta_{t-1}-\theta_{t}$ are 1.88$e^{-6}$ for our method and 6.90$e^{-6}$ for the \nj{other}.

\vspace{-2mm}
\section{Conclusion}
\vspace{-2mm}
\hj{In this paper, we have addressed and tackled the scarcity of sequential RGB dataset which limits conventional methods from exploiting temporal image
features for 3D HPE.}
We have proposed a novel method to generate SeqHAND dataset, a dataset with sequential RGB image frames of synthetic hand poses in motions that are interpolated from existing static pose annotations.
We then proposed a framework that exploits visuo-temporal features for 3D hand pose estimations in a recurrent manner.
We have implemented a cost function considering the temporal smoothness of sequential hand pose estimations.
Our proposed method outperforms other existing approaches that take RGB-only inputs that are based on solely appearance-based methods, and consequently produces pose-flow estimations that mimick natural movements of human hands.

\hj{We also plan to enable the framework to solve (self-)occlusion problems more robustly. With sequential inputs, we were able to witness possibility of overcoming conventional struggle against occlusion problems in the literature of 3D hand pose estimations. }

\newpage
\subsubsection{Acknowledgement}
 This work was supported by IITP grant funded by the Korea government (MSIT) (No.2019-0-01367, Babymind) and Next-Generation Information Computing Development Program through the NRF of Korea (2017M3C4A7077582).

%
%
\bibliographystyle{splncs04}
\bibliography{egbib}
\end{document}